\documentclass[pdflatex,sn-mathphys-num]{sn-jnl}% Math and Physical Sciences Numbered Reference Style
\usepackage{graphicx}%
\usepackage{multirow}%
\usepackage{amsmath,amssymb,amsfonts}%
\usepackage{amsthm}%
\usepackage{mathrsfs}%
\usepackage[title]{appendix}%
\usepackage{xcolor}%
\usepackage{textcomp}%
\usepackage{manyfoot}%
\usepackage{booktabs}%
\usepackage{algorithm}%
\usepackage{algorithmicx}%
\usepackage{algpseudocode}%
\usepackage{listings}%
\usepackage{subcaption}%
\theoremstyle{thmstyleone}%
\theoremstyle{thmstyletwo}%

\theoremstyle{thmstylethree}%

\begin{document}

%\title[Article Title]{Auto-rickshaw Detection using Machine Learning}
\title[Article Title]{Detecting Unauthorized Vehicles using Deep Learning for Smart Cities: A Case Study on Bangladesh}


\author[1]{\fnm{Sudipto Das} \sur{Sukanto}}\email{sukantodas302@gmail.com}
\author[1]{\fnm{Diponker} \sur{Roy}}\email{dibbyoroy7@gmail.com}
\author[1]{\fnm{Fahim} \sur{Shakil}}\email{littlefahim2000@gmail.com}
\author[1]{\fnm{Nirjhar} \sur{Singha}}\email{nirjharsingha@gmail.com}
\author[1]{\fnm{Abdullah} \sur{Asik}}\email{ashikcse2732@gmail.com}
\author[1]{\fnm{Aniket} \sur{Joarder}}\email{joarderaniket@gmail.com}

% Corresponding authors
\author[2]{\fnm{Mridha Md. Nafis} \sur{Fuad}}\email{fuad@iit.du.ac.bd}
\author*[1]{\fnm{Muhammad} \sur{Ibrahim}}\email{ibrahim313@du.ac.bd}

\affil[1]{\orgdiv{Department of Computer Science \& Engineering}, 
           \orgname{University of Dhaka}, 
           \orgaddress{\city{Dhaka}, \country{Bangladesh}}}
\affil[2]{\orgdiv{Institute of Information Technology}, 
           \orgname{University of Dhaka}, 
           \orgaddress{\city{Dhaka}, \country{Bangladesh}}}


\abstract{Modes of transportation vary across countries depending on geographical location and cultural context. In South Asian countries rickshaws are among the most common means of local transport. Based on their mode of operation, rickshaws in cities across Bangladesh can be broadly classified into non-auto (pedal-powered) and auto-rickshaws (motorized). Monitoring the movement of auto-rickshaws is necessary as traffic rules often restrict auto-rickshaws from accessing certain routes. However, existing surveillance systems make it quite difficult to monitor them due to their similarity to other vehicles, especially non-auto rickshaws whereas manual video analysis is too time-consuming. This paper presents a machine learning-based approach to automatically detect auto-rickshaws in traffic images. In this system, we used real-time object detection using the YOLOv8 model. For training purposes, we prepared a set of 1,730 annotated images that were captured under various traffic conditions. The results show that our proposed model performs well in real-time auto-rickshaw detection and offers an mAP50 of 83.447\% and binary precision and recall values above 78\%, demonstrating its effectiveness in handling both dense and sparse traffic scenarios. Our dataset has been publicly released for further research.}

%%================================%%
%% Sample for structured abstract %%
%%================================%%




\maketitle

\section{Introduction}\label{sec1}

In developing countries, battery-operated auto-rickshaws are an important mode of transportation that offers citizens the ability to travel in urban areas at an affordable price. However, they pose a significant threat to road safety on certain roads in cities. Manual monitoring of their movement on different roads and highways in cities using traditional surveillance methods is inefficient. With the rise of smart city initiatives, automated traffic surveillance systems are replacing traditional surveillance methods, making it much easier to enforce traffic rules. In real-time traffic decisions, automatically detecting specific vehicle types (e.g., auto-rickshaw) can play a crucial role in maintaining traffic regulations. This paper proposes a method that uses real-time object detection using deep learning methods to detect auto-rickshaws in traffic images or videos of CCTV cameras at traffic signals. Our method using real-time object detection with the YOLOv8 framework addresses common urban surveillance challenges and aims to provide a scalable tool for intelligent traffic monitoring. 

Image-based classification with machine learning algorithms has long been a central theme in computer vision \cite{rehana2023plant}, \cite{tahsin2025paddyvarietybd}, \cite{ferdaus2025mangoimagebd}, \cite{hasan2024selecting}. Significant progress has been made in autonomous driving, satellite monitoring, and remote sensing applications. Urban analytics has also been leveraging from machine learning algorithms \cite{rasel2025road}, \cite{Lubaina2024FootpathVision}. However, very few, if at all, no dataset, to the best of our knowledge, focuses specifically on detecting unauthorized vehicles in dense traffic urban areas.

Recent advances in deep learning have led to a revolution in computer vision. Models like CNNs excel at image classification, and object detectors like YOLO and SSD have enabled us to localize objects in real-time. In the field of intelligent transportation systems, these tools are widely used for tasks such as vehicle counting, license plate recognition, traffic monitoring, etc. However, most of the existing models are trained on datasets like \textbf{COCO} or \textbf{KITTI}, which, despite being globally recognized datasets, lack region-specific vehicles, such as battery-operated auto-rickshaws. These auto-rickshaws are very common in Bangladesh, but not seen in many other countries, especially developed countries of Europe and America. As a result, in cities in Bangladesh and similar countries, where traffic is dense and chaotic, the effectiveness of these datasets is limited by the discrepancy between training and validation data distribution. This further proves the necessity of domain-specific datasets tailored to detect auto-rickshaws in real-world traffic. 

Auto-rickshaws can vary greatly in appearance due to different trends in different cities and are often seen in scenes with different lighting. To handle all these variations efficiently, our system used a dataset containing a large number of images of auto-rickshaws of different appearances that were captured in different lighting conditions. 

Unlike existing vehicle detection systems, our system treats auto-rickshaws as a separate class. In the absence of public datasets tailored for auto-rickshaws, we built our own dataset, collecting images from various cities under various lighting, traffic, and daytime conditions. We benchmark this dataset using models like YOLO for fast and efficient object detection. Through a dedicated, standard, ready-to-use dataset and detection pipeline, this system offers a practical and region-aware solution to automate traffic monitoring tasks. The model trained with the newly annotated dataset achieves an mAP50 score of 83.447\% and binary precision and recall values above 78\%.  Our developed dataset has been publicly released for further research (\url{https://data.mendeley.com/datasets/bg6wvvhsjh/1}).

\section{Related Work}\label{sec:relatedwork}
Over the past decade, the field of object detection has witnessed rapid progress, largely due to advanced model architectures, efficient loss functions and training strategies. Unlike earlier approaches that may fail to balance between accuracy and real-time efficiency, modern systems are trying to close this gap, making these models suitable for real-life applications like intelligent traffic surveillance and transportation.

Rehana et al.(2023) \cite{rehana2023plant} proposed a lightweight modified R-CNN model for early detection of tomato leaf diseases, optimizing computational efficiency and enabling drone-based agricultural surveillance for automated crop health monitoring. Tahsin et al. (2025) \cite{tahsin2025paddyvarietybd} introduced PaddyVarietyBD, a large-scale dataset of over 14,000 microscopic rice kernel images from Bangladeshi research institutes to facilitate varietal classification and AI-driven agricultural analysis. Ferdaus et al. (2025) \cite{ferdaus2025mangoimagebd} developed MangoImageBD, a 28,515-image dataset of fifteen Bangladeshi mango varieties designed for machine learning tasks such as classification, detection, and segmentation to advance precision horticulture and quality assessment. Hasan et al. (2024) \cite{hasan2024selecting} presented a genetic algorithm-based method for adaptive layer selection in CNN transfer learning, reducing training time and parameters while maintaining high accuracy across datasets like Food-101 and MangoLeafBD. Hossen et al. (2025) \cite{rasel2025road} designed a YOLOv9-based model using polygonal annotations for accurate detection of road damages and manholes in Dhaka, achieving strong performance and supporting scalable smart city infrastructure monitoring.

Lubaina et al. (2024) \cite{Lubaina2024FootpathVision} introduced FootpathVision, a comprehensive image dataset with deep learning baselines for detecting footpath encroachments, supporting urban infrastructure management and smart city research. Fahim (2024) \cite{fahim2024finetuning} fine-tuned the YOLOv9 model for vehicle detection in Dhaka’s traffic environments, demonstrating its applicability to intelligent transportation systems in developing cities. Wang et al. (2020) \cite{wang2020robust} proposed a robust vehicle detection framework for smart city traffic surveillance using advanced deep learning techniques to enhance accuracy and reliability in complex urban conditions.

Ren et al. (2015) \cite{ren2015faster} introduced Faster R-CNN, a unified object detection framework integrating Region Proposal Networks with Fast R-CNN to generate efficient in-network proposals, achieving higher accuracy and lower computational cost but remaining unsuitable for real-time applications due to its two-stage design. Lin et al. (2017) \cite{lin2017retinanet} proposed RetinaNet, a one-stage detector leveraging the novel Focal Loss function to address class imbalance and achieve high accuracy with faster inference, making it ideal for real-time detection tasks. Carion et al. (2020) \cite{carion2020detr} presented DETR, a transformer-based object detector eliminating anchor boxes and proposals through self-attention and the Hungarian algorithm, offering competitive accuracy but limited by slow convergence and high computational demand. Ultralytics (2023) \cite{ultralytics2023yolov8} released YOLOv8, an advanced and flexible evolution of the YOLO family supporting detection, segmentation, and classification with enhanced multi-scale representation and adaptability for domain-specific applications.

Autogyro’s YOLOv8 (2023) \cite{autogyro2023yolov8}, forked from Ultralytics, offers a fast and flexible framework for
object detection, segmentation, and classification using PyTorch. This
open-source implementation served as the base for our object detection experiments. Shohan et al. (2025) \cite{shohan2025data} introduced a data-driven machine learning approach to predict crime
occurrence in Bangladesh. By combining crime, geographic, temporal, and demographic data,
they achieved improved prediction accuracy providing a solid baseline for regional
crime prediction and analysis. Mahalakshmi et al. (2023) \cite{mahalakshmi2023vehicle} developed a hybrid vehicle-detection model combining CNN and
YOLOv3, achieving strong real-time performance on traffic datasets. Husain et al. (2025) \cite{husain2025moving} focused on vehicle detection and tracking in hazy weather conditions using
enhanced vision-based algorithms. Their method addressed visibility degradation in traffic
footage, contributing to more reliable surveillance under adverse environments. Shepelev et al. (2023) \cite{shepelev2023using} utilized computer vision to study vehicle sequencing at regulated
intersections. Their system analyzed video data to detect and track vehicles, offering insights
into flow efficiency and intersection performance.

El-Alami et al. (2023) \cite{el2023efficient} proposed an efficient hybrid model that integrates background subtraction
with deep learning for vehicle detection and tracking. The system effectively reduces false
positives while enhancing stability in both day and night traffic scenes, making it useful for
continuous urban monitoring.
El-Alami et al. (2024) \cite{el2024review} reviewed a variety of object detection techniques applied in traffic
surveillance systems. They discussed the evolution from traditional feature-based methods to
modern CNN-based approaches like YOLO and SSD, while emphasizing improvements in
accuracy, efficiency, and adaptability to real-world traffic environments.
Vikruthi et al. (2025) \cite{vikruthi2025detection} developed a deep learning-based system combining image processing and neural networks to detect emergency vehicles in real time, enabling traffic-free routing and improved smart city emergency response. Ghoniem et al. (2021) \cite{ghoniem2021intelligent} performed a systematic review of AI- and IoT-enabled intelligent surveillance systems for smart cities, emphasizing their role in enhancing traffic management and urban safety. Aeri and Purohit (2024) \cite{aeri2024unveiling} analyzed AI-driven vehicle tracking and detection frameworks such as YOLO and Faster R-CNN, demonstrating their effectiveness in managing challenges like occlusion, motion blur, and lighting variability in real-world traffic environments.

Karri et al. (2024) \cite{karri2024recent} presented a comprehensive review on technological advancements in smart
city management. Their work covered recent trends in data-driven solutions, IoT integration, and
intelligent infrastructure systems that aim to enhance urban efficiency and sustainability.
Kiran et al. (2022) \cite{kiran2022edge} proposed a noise-robust deep learning network designed for vehicle
classification. Their edge-preserving model effectively minimizes distortion caused by
environmental noise, leading to more accurate vehicle recognition under challenging real-world
conditions. Mohandoss and Rangaraj (2024) \cite{mohandoss2024performance} analyzed surveillance video object detection using the Lunet
algorithm. Their study focused on improving detection speed and reliability in traffic video feeds,
demonstrating the model’s capability for efficient real-time monitoring applications.
Ghahremannezhad et al. (2023) \cite{ghahremannezhad2023object} provided a detailed survey on object detection in traffic videos.
They compared multiple detection frameworks and deep learning methods, highlighting current
challenges and trends in intelligent transportation systems research. Kiran et al. (2024) \cite{kiran2024vehicle} investigated vehicle detection performance in various weather conditions
using an enhanced YOLO model with complex wavelet transformation. Their approach
improved feature extraction and detection consistency across diverse environmental settings.

\section{Methods}
The process is structured into two primary phases: (i) data collection and preprocessing, and (ii) experimental setup and evaluation. The implementation was carried out using Python, incorporating the Ultralytics YOLOv8 framework for object detection. A modular architecture was adopted to ensure a systematic and efficient workflow throughout the entire detection pipeline.

\subsection{Data Collection and Preprocessing}
Manual data collection and annotation is done as global datasets do not include images of auto-rickshaws to differentiate from pedal powered (non-auto) rickshaws. Diverse real-world traffic photos were collected by hand and carefully labeled as part of building a computer vision model that works well in domain-specific scenario. 

\subsubsection{Data Collection}

Smartphone cameras and portable cameras were used to take high-resolution pictures for our dataset. The images were collected from diverse sources like - (i) live traffic junctions and intersections, (ii) roadside observations near public places (such as terminals and busy markets), and (iii) still frames of street surveillance-style recordings. Pictures were taken at different times of the day and night to cover a variety of lighting conditions and simulate the real-world environments. Daytime images were taken in the presence of sufficient natural sunlight to show the entire color spectrum and help the model recognize the shape, paint patterns and other structural and aesthetic characteristics of auto-rickshaws. Nighttime images were taken to enhance detection in practical scenarios of dim backgrounds and poor visibility.

\begin{figure}[t]
    \centering
    
    % First row (3 images)
    \begin{subfigure}{0.32\textwidth}
        \centering
        \includegraphics[width=\linewidth,height=4.2cm]{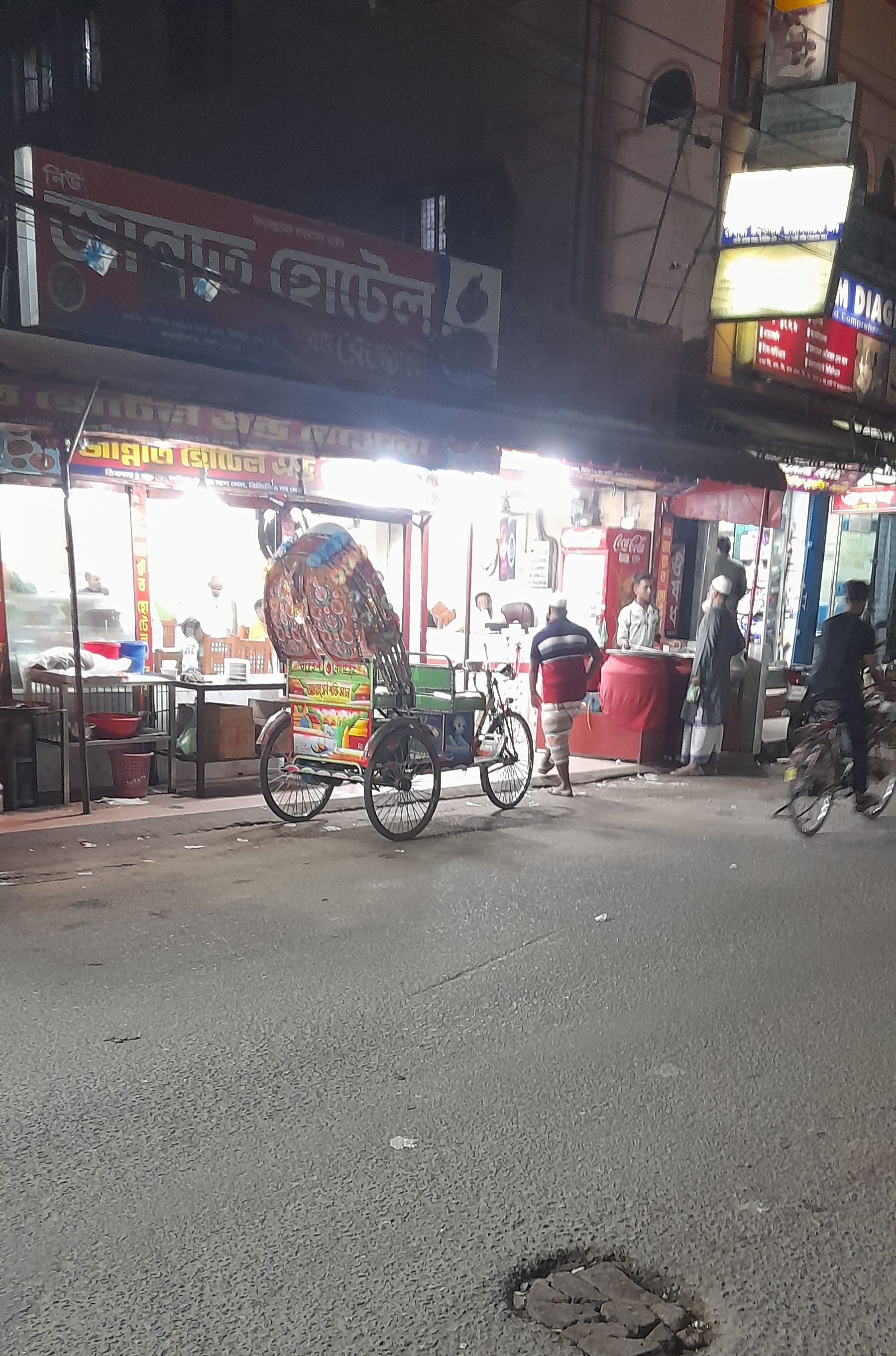}
        \label{fig:recall_curve}
    \end{subfigure}
    \hfill
    \begin{subfigure}{0.32\textwidth}
        \centering
        \includegraphics[width=\linewidth,height=4.2cm]{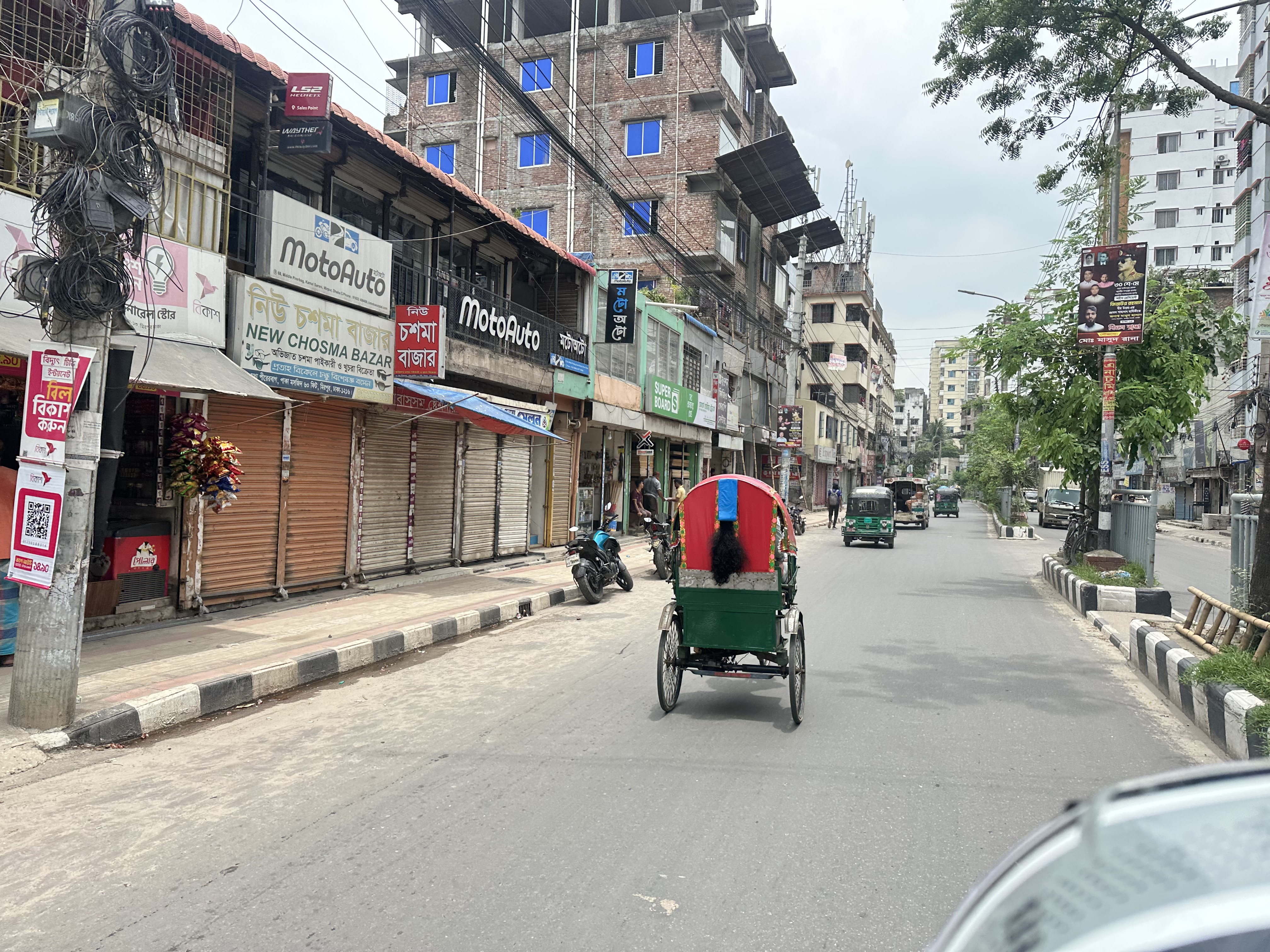}
        \label{fig:f1_curve_1}
    \end{subfigure}
    \hfill
    \begin{subfigure}{0.32\textwidth}
        \centering
        \includegraphics[width=\linewidth,height=4.2cm]{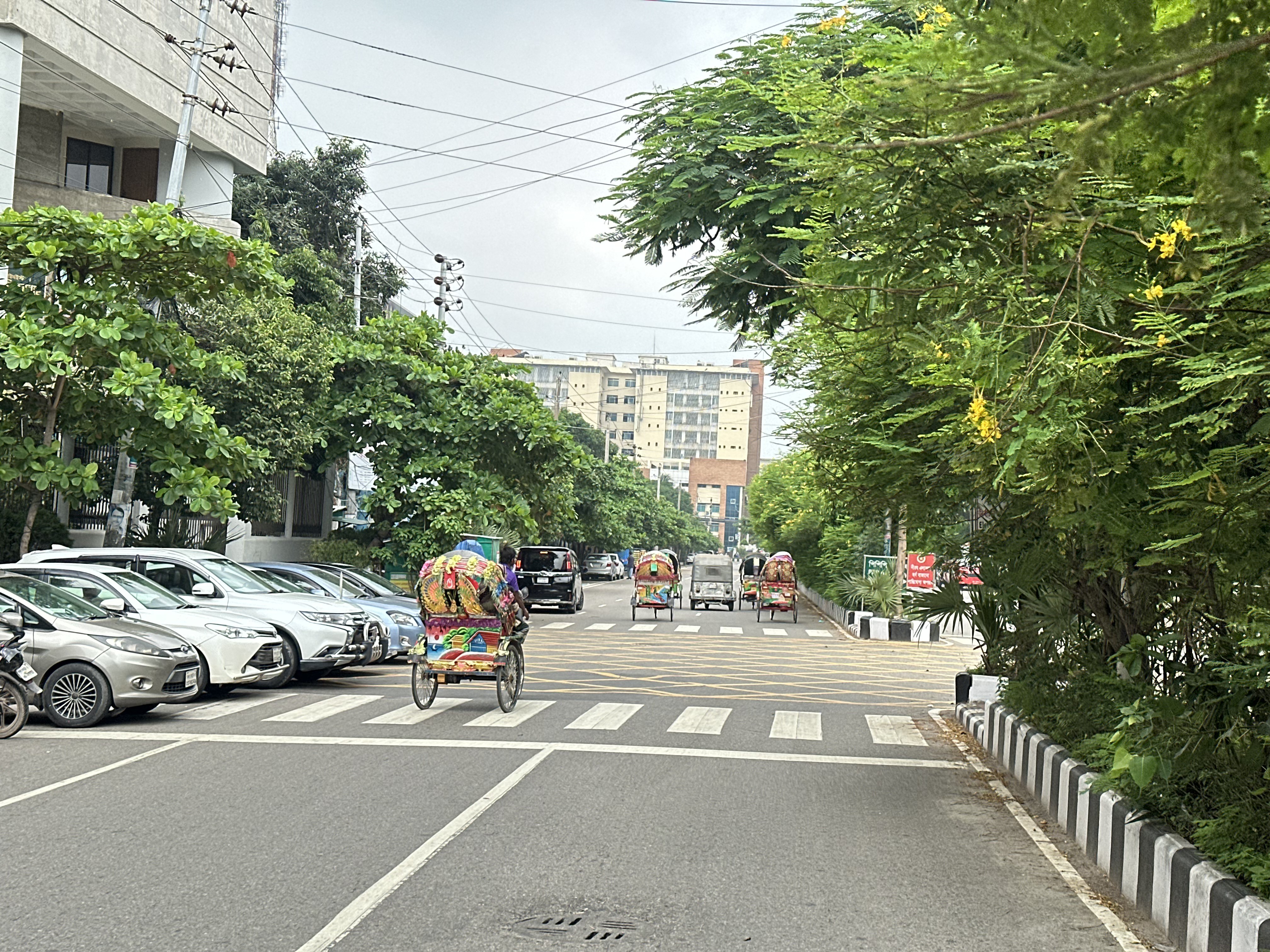}
        \label{fig:precision_curve}
    \end{subfigure}
    
    \vspace{1em} % Space between rows
    
    % Second row (3 images)
    \begin{subfigure}{0.32\textwidth}
        \centering
        \includegraphics[width=\linewidth,height=4.2cm]{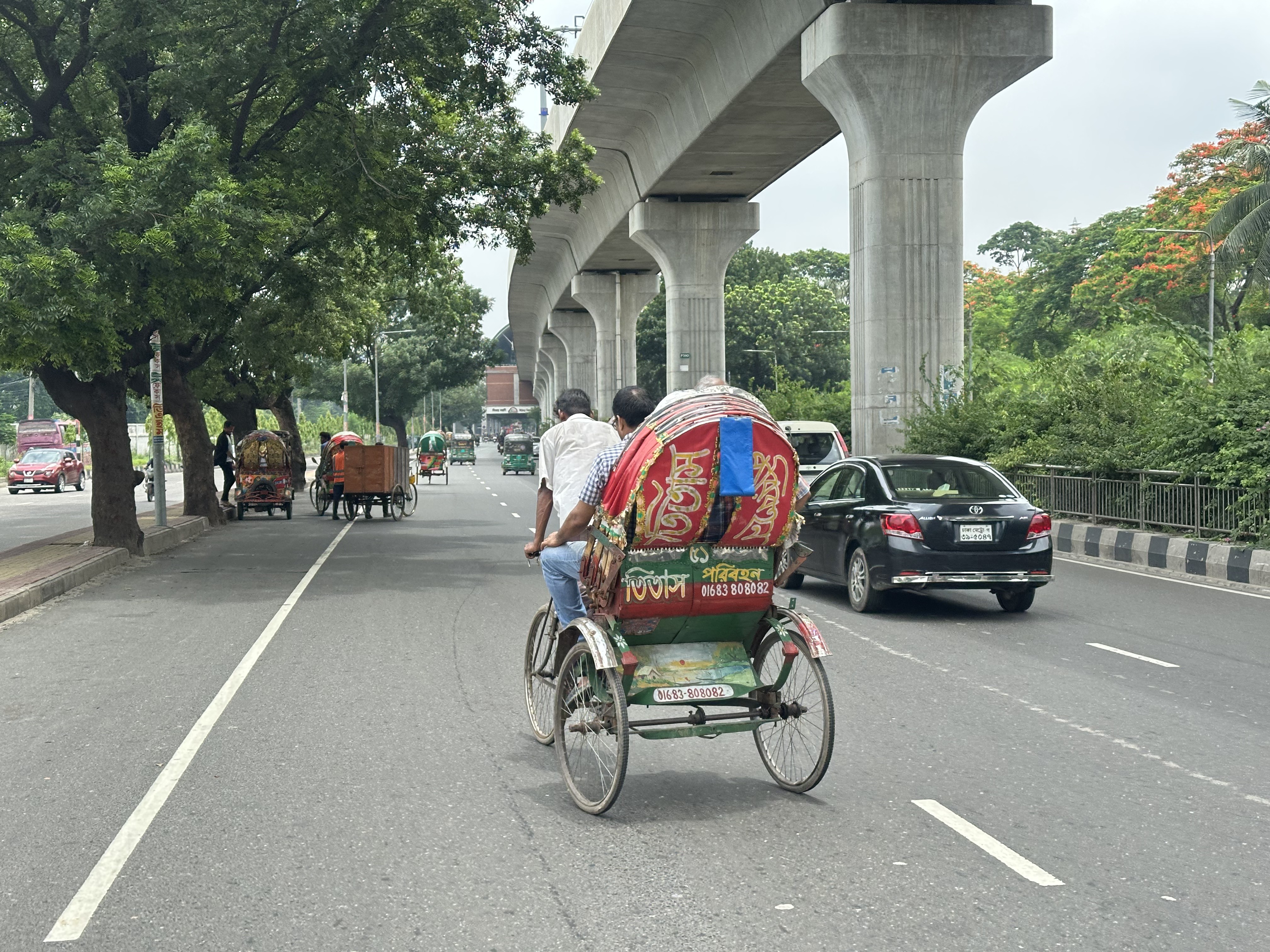}
        \label{fig:pr_curve}
    \end{subfigure}
    \hfill
    \begin{subfigure}{0.32\textwidth}
        \centering
        \includegraphics[width=\linewidth,height=4.2cm]{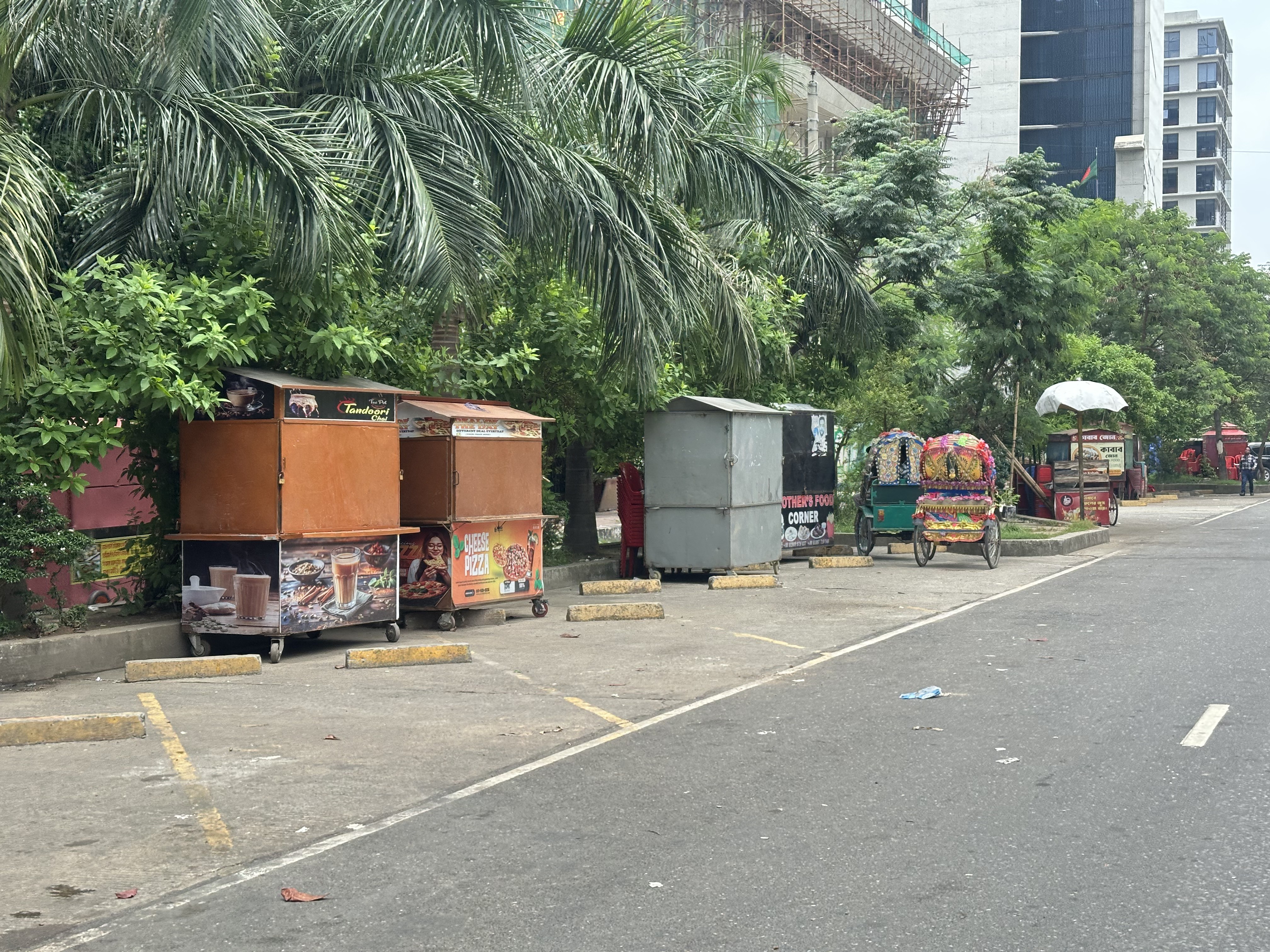}
        \label{fig:f1_curve_2}
    \end{subfigure}
    \hfill
    \begin{subfigure}{0.32\textwidth}
        \centering
        \includegraphics[width=\linewidth,height=4.2cm]{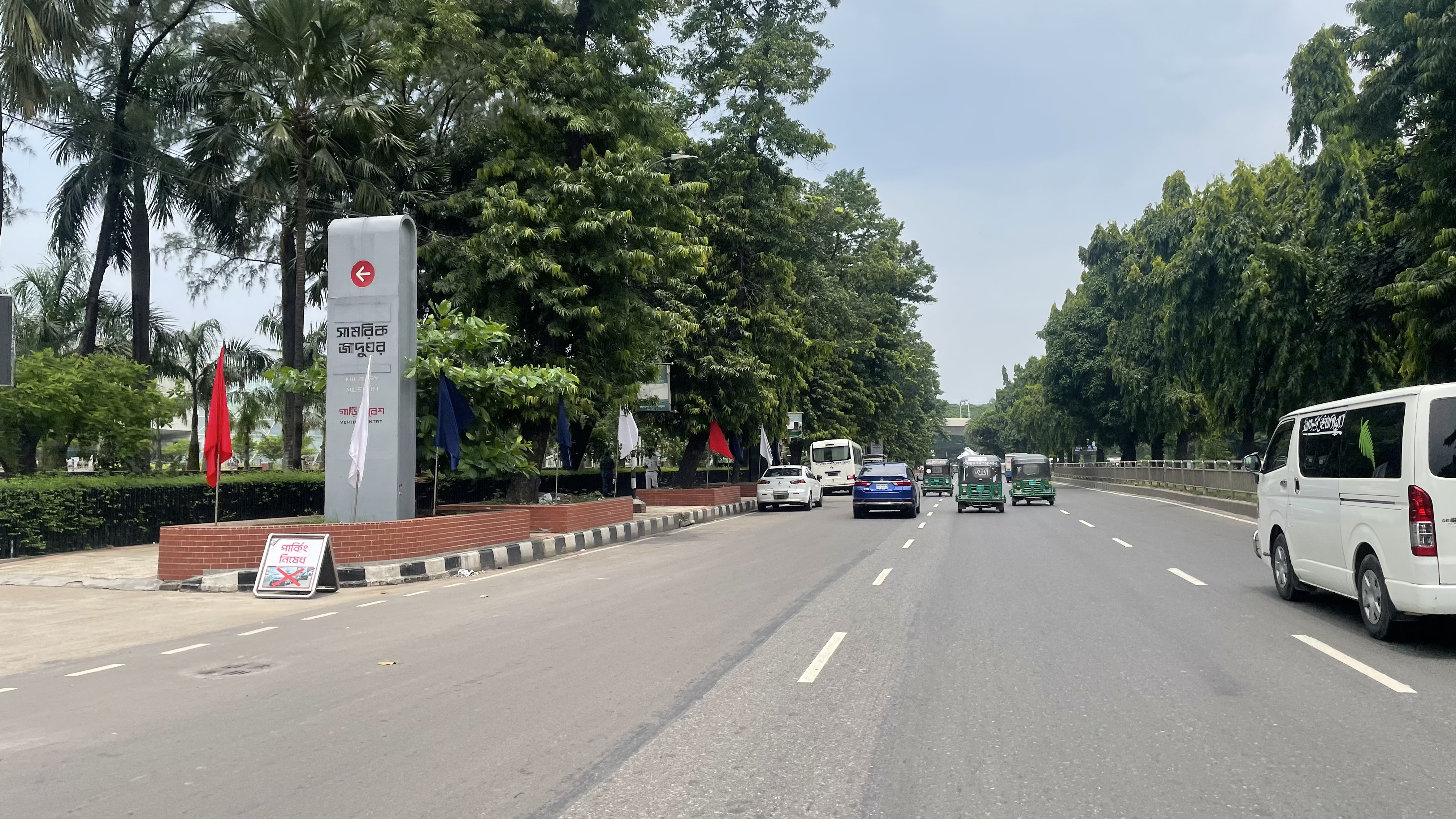}
        \label{fig:f1_curve_3}
    \end{subfigure}
    
    \caption{Sample images without bounding box}
    \label{fig:raw_images}
\end{figure}

The images were collected from a wide range of areas of Dhaka and nearby cities, including Narayanganj, Gazipur, and Savar. Dhaka being the most crowded city is prioritized for image collection to reduce challenges in real-world traffic scenarios. These images include scenes with a variety of congested traffic, such as cars, rickshaws, buses, and even people. We also collected images from other divisional cities like- Cumilla, Khulna, etc. to represent less crowded scenarios in our dataset. These images are needed to distinguish between auto-rickshaws and stationary background items such as poles, road dividers, or parked vehicles. Some images show little or no vehicles at all. Moreover, this geographical diversity is crucial as auto-rickshaws differ significantly in design and color schemes in different regions. For example, in central Dhaka, auto-rickshaws are generally brightly painted and may also carry commercial banners or logos. In suburban areas, auto-rickshaws with simple designs are mostly seen. Some even have faded or unpainted body parts. This broad coverage ensures that the trained model does not overfit the visual patterns from a single location and is adaptive when facing various real-life challenges. Therefore, the images were captured in different scenarios.

In order to ensure class balance so that the dataset does not get skewed, our dataset follows the given distribution.
\begin{itemize}
    \item \textbf{Auto-rickshaws Present:} Approximately 60–63\% of the images include one or more auto-rickshaws.
    \item \textbf{Non-Auto Vehicles Only:} Around 35-37\% of the images include non-auto rickshaws along with other types of vehicles (cars, vans, trucks, motorcycles) without any auto-rickshaws.
    \item \textbf{Empty Backgrounds:} The remaining 2-3\% are background-only images with no vehicles. These images act as negative samples to enhance the discriminative power of the model.
\end{itemize}

\begin{figure}[t]
    \centering
    
    % First row (3 images)
    \begin{subfigure}{0.32\textwidth}
        \centering
        \includegraphics[width=\linewidth,height=4.2cm]{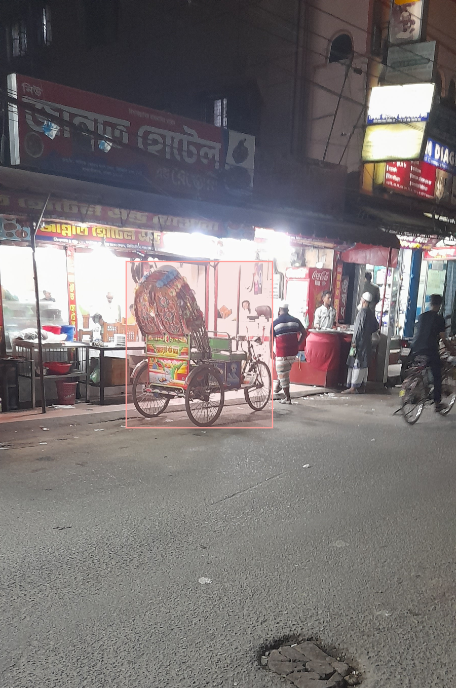}
        \label{fig:recall_curve}
    \end{subfigure}
    \hfill
    \begin{subfigure}{0.32\textwidth}
        \centering
        \includegraphics[width=\linewidth,height=4.2cm]{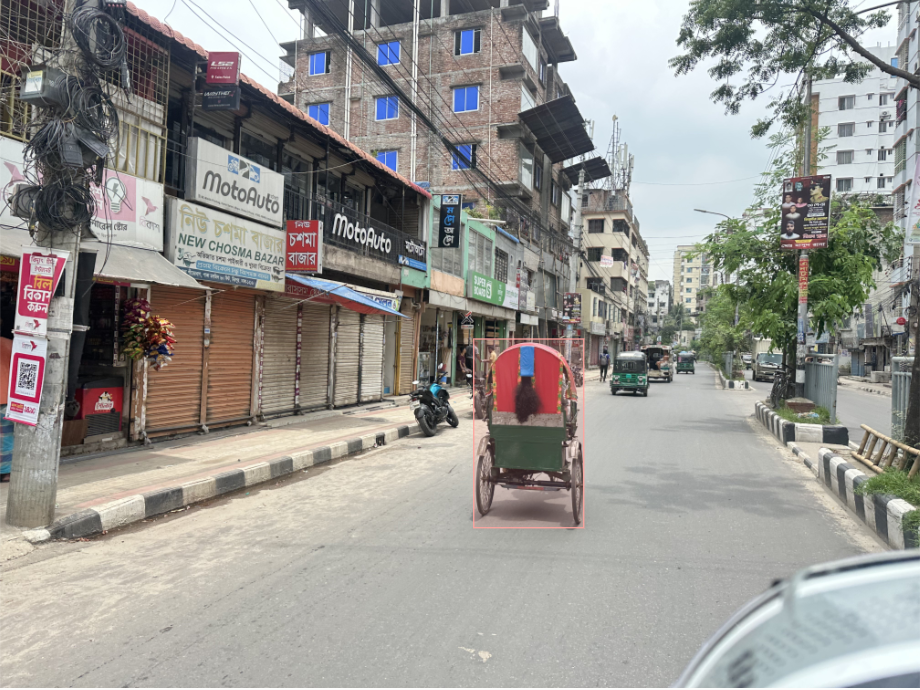}
        \label{fig:f1_curve_1}
    \end{subfigure}
    \hfill
    \begin{subfigure}{0.32\textwidth}
        \centering
        \includegraphics[width=\linewidth,height=4.2cm]{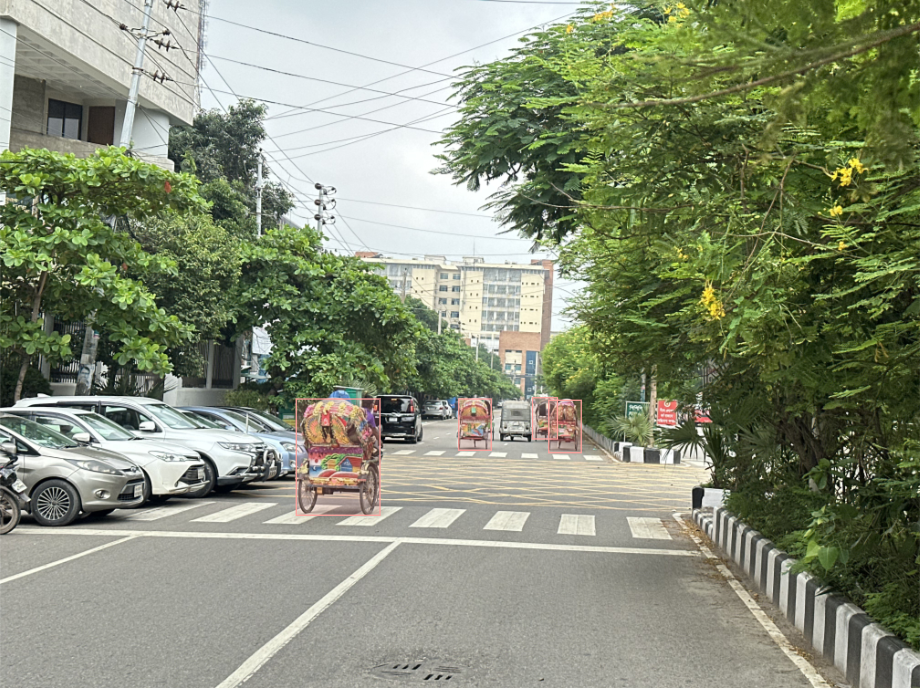}
        \label{fig:precision_curve}
    \end{subfigure}
    
    \vspace{1em} % Space between rows
    
    % Second row (3 images)
    \begin{subfigure}{0.32\textwidth}
        \centering
        \includegraphics[width=\linewidth,height=4.2cm]{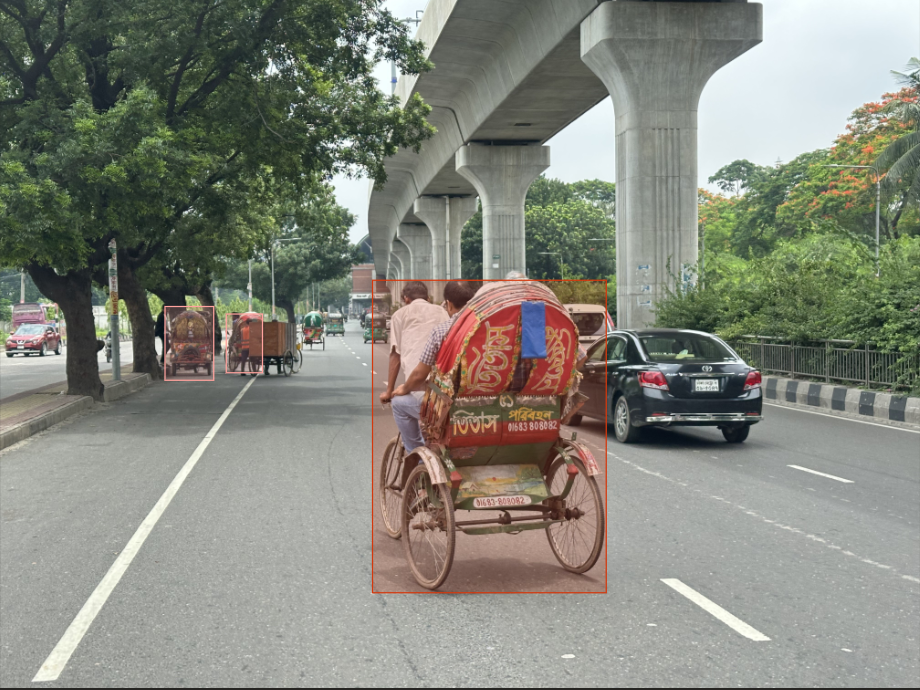}
        \label{fig:pr_curve}
    \end{subfigure}
    \hfill
    \begin{subfigure}{0.32\textwidth}
        \centering
        \includegraphics[width=\linewidth,height=4.2cm]{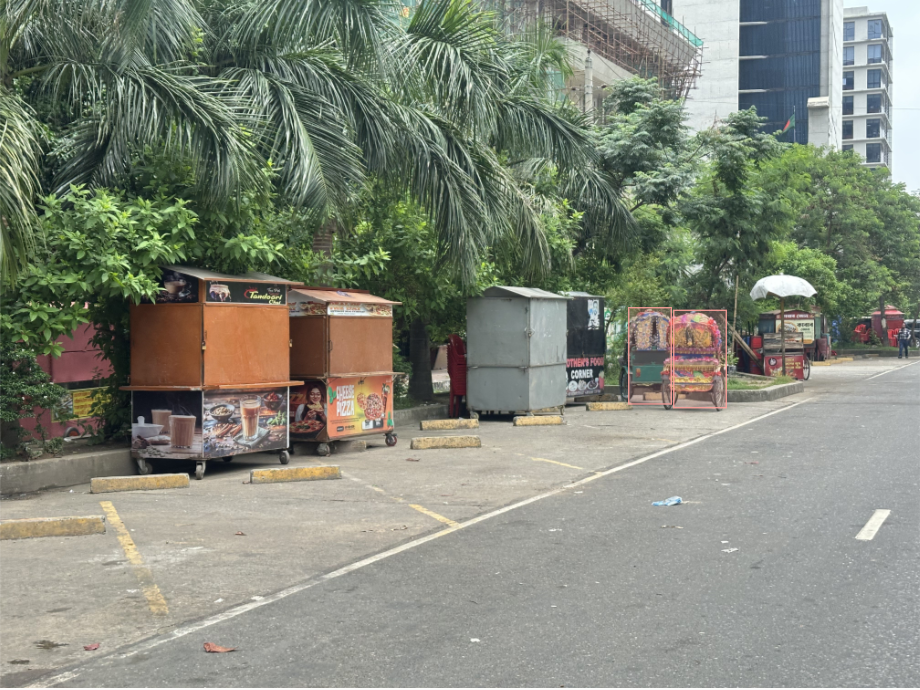}
        \label{fig:f1_curve_2}
    \end{subfigure}
    \hfill
    \begin{subfigure}{0.32\textwidth}
        \centering
        \includegraphics[width=\linewidth,height=4.2cm]{no_rick.jpeg}
        \label{fig:f1_curve_3}
    \end{subfigure}
    
    \caption{Sample images with bounding box}
    \label{fig:annotated_images}
\end{figure}

As auto-rickshaws make up only a smaller percentage of the total traffic volume in the real world, this balanced method was chosen to avoid bias and overfitting. Fig. \ref{fig:raw_images} and Fig. \ref{fig:annotated_images} shows sample raw images collected and their corresponding annotated bounding boxes marking auto-rickshaws respectively.

\subsubsection{Data Preprocessing}

The collected 1331 images were manually annotated using the open-source tool Label Studio \footnote{https://labelstud.io/}. The annotations were done following the standard YOLO object detection format. We marked each visible auto-rickshaw instance using a rectangular bounding box indicating one class. Non-auto rickshaws were similarly annotated, indicating another class. For quality assurance each bounding box was carefully drawn to cover the auto or non-auto rickshaw only, avoiding overlaps with other vehicles. Difficult instances, such as truncated views, were labeled if they were visually identifiable. Some images that are not visually identifiable or too difficult to annotate were filtered from the dataset.

Each annotated image has a corresponding label file, which stores the coordinates of the bounding box and class names. The complete dataset was divided into two disjoint subsets: 90\% training and 10\% validation. Algorithm \ref{dataset-preparation} represents the steps taken to prepare the data for the training and validation step. We carefully ensured that each subset included a mix of images representing different environments, lighting conditions, and vehicle types. We also maintained the class balance for the dataset so that it would not get skewed. This setup allowed us to determine the ability of the model to evaluate unseen situations and adapt to changing scenarios. 
This extensive data preparation process was tailored to detect auto-rickshaws in the urban environments of Bangladesh, incorporating a wide range of real-world situations.

% \begin{algorithm}
% \caption{Dataset Preparation and Train/Validation Split}
% \begin{algorithmic}[1]
% \State \textbf{Input:} Labeled images directory, train\_ratio = 0.9
% \State \textbf{Output:} YOLO-formatted dataset with train/val split
% \State
% \State Load all image files from source directory
% \State Shuffle image list randomly
% \State Calculate split\_index = total\_images $\times$ train\_ratio
% \State Split images into train\_files and val\_files
% \State
% \For{each image in train\_files}
%     \State Copy image to train directory
%     \State Copy corresponding label file to train/labels
% \EndFor
% \For{each image in val\_files}
%     \State Copy image to val directory
%     \State Copy corresponding label file to val/labels
% \EndFor
% \State Generate YAML configuration file
% \State Validate dataset integrity
% \end{algorithmic}
% \end{algorithm}

\begin{algorithm}[H]
\begin{algorithmic}[1]
\State \textbf{Input:} Labeled images directory, train\_ratio = 0.8
\State \textbf{Output:} YOLO-formatted dataset with train/validation split
\State

\State \textbf{Step 1:} Load all image files from the source directory.  
\State \textbf{Step 2:} Shuffle the image list randomly.  
\State \textbf{Step 3:} Calculate the split index as: split\_index = total\_images $\times$ train\_ratio.  
\State \textbf{Step 4:} Divide the dataset into two sets: train\_files and val\_files.  

\State \textbf{Step 5:} For each image in train\_files:\\  
\hspace{1em} a) Copy the image to the train directory.\\  
\hspace{1em} b) Copy the corresponding label file to train/labels.  

\State \textbf{Step 6:} For each image in val\_files:\\  
\hspace{1em} a) Copy the image to the val directory.\\ 
\hspace{1em} b) Copy the corresponding label file to val/labels.  

\State \textbf{Step 7:} Generate the YAML configuration file.  
\State \textbf{Step 8:} Validate the dataset integrity.  

\end{algorithmic}
\caption{Dataset Preparation and Train/Validation Split}
\label{dataset-preparation}
\end{algorithm}

\subsection{Experimental Setup and Evaluation}

The model was trained using the default Ultralytics training pipeline with built-in optimizer and loss functions. 

\paragraph{Hardware Configuration:}
The model was trained on a Mac Mini with the Apple M4 chip, 16GB of unified memory, and 256GB of storage. The total computational training time required is 6 hours and the training configuration is shown in Table \ref{table:training-configuration}.

\begin{table}[t]
\centering
\caption{Training Configuration of the YOLOv8n Model}
\label{table:training-configuration}
\begin{tabular}{ll}
\toprule
\textbf{Parameter} & \textbf{Value / Description} \\
\midrule
Base Model & YOLOv8n (nano), pre-trained on \texttt{yolov8n.pt} \\
Class Labels & 2 (auto, non-auto rickshaw) \\
Total Images & 1,331 \\
Image Size & 640$\times$640 pixels \\
Train/Validation Split & 90\% / 10\% \\
Epochs & 50 \\
Batch Size & 8 \\
Confidence Threshold & 0.25 \\
Bounding Boxes & Precise coordinates for each detection \\
Early Stopping & Disabled (\texttt{patience = 0}) \\
Workers & 8 (for parallel data loading) \\
Device & CPU (automatically selected) \\
Output Formats & JSON results and annotated images \\
Image Formats & JPG, JPEG, PNG \\
Label Format & YOLO format (normalized coordinates) \\
Test Set & Validation set used for evaluation \\
\bottomrule
\end{tabular}
\end{table}

\paragraph{Dataset Information:}
\begin{itemize}
    \item \textbf{Annotation Format:} YOLO-compatible bounding box annotations
    \item \textbf{Data Augmentation:} Default YOLOv8 augmentations applied automatically (not explicitly configured)
\end{itemize}

The trained model generates a list of bounding boxes, class labels, and confidence scores per image as output. We defined the confidence threshold at 0.25 to refine the detection results and eliminate overlapping predictions. With a comprehensive pipeline incorporating systematic data partitioning, training, and hyperparameter tuning, our designed system can assess unseen data of real-world scenarios with a higher confidence rate.

For evaluating the performance of the object detector, we used \textbf{Mean Average Precision (mAP)}, which acts as a measure of the model's precision and recall across different thresholds and objects. This is a widely accepted metric in the field of object detection to assess the detection accuracy and localization precision of a model.

\subsubsection{Training and Hyperparameter Tuning}

Multiple training sessions were conducted to fine-tune the model. Key hyperparameters such as batch size, learning rate, and number of epochs were iteratively adjusted to improve performance. During the training process, we ensured that we reduced overfitting as much as possible. Apart from the built-in data augmentation the YOLOv8 model provides, we did not apply data augmentation actively since we already had sufficient images for training and validation purposes.

\subsubsection{Detection and Inference Algorithm}

The detection system implements a real-time inference pipeline using the trained YOLOv8 model. Algorithm \ref{algo_detection} demonstrates the core detection algorithm processes from taking input images to returning bounding boxes with confidence scores for each detected rickshaw.

\subsubsection{Validation and Qualitative Assessment}
We tried to replicate real-world deployment scenarios during the training and validation phases. In our evaluation, we paid particular attention to detecting auto-rickshaws in low-visibility conditions, such as at night or under shadows, differentiating them from visually similar vehicles, including non-auto rickshaws and vans, and minimizing false positives in images containing only background without any vehicles.

% \begin{algorithm}[H]
% \caption{Rickshaw Detection and Inference}
% \begin{algorithmic}[1]
% \State \textbf{Input:} Image path, confidence\_threshold = 0.25
% \State \textbf{Output:} Detected rickshaws with bounding boxes and confidence scores
% \State
% \State Load trained YOLOv8 model
% \State Read and preprocess input image
% \State Run YOLOv8 prediction with confidence thresholding
% \State
% \For{each detection in results}
%     \State Extract bounding box coordinates (x1, y1, x2, y2)
%     \State Get class ID and confidence score
%     \State Map class ID to class name ('auto' or 'non-auto')
%     \State Draw colored bounding box (green for auto, red for non-auto)
%     \State Add confidence label to bounding box
%     \State Store detection in results list
% \EndFor
% \State Return annotated image and detection list
% \end{algorithmic}
% \end{algorithm}

\begin{algorithm}[H]
\caption{Rickshaw Detection and Inference}
\begin{algorithmic}[1]
\State \textbf{Input:} Image path, confidence\_threshold = 0.25  
\State \textbf{Output:} Detected rickshaws with bounding boxes and confidence scores  
\State

\State \textbf{Step 1:} Load the trained YOLOv8 model.  
\State \textbf{Step 2:} Read and preprocess the input image.  
\State \textbf{Step 3:} Run YOLOv8 prediction with the given confidence threshold.  

\State \textbf{Step 4:} For each detection in the results:\\  
\hspace{1em} a) Extract bounding box coordinates $(x1, y1, x2, y2)$.\\  
\hspace{1em} b) Get the class ID and corresponding confidence score.\\  
\hspace{1em} c) Map the class ID to its name (\texttt{auto} or \texttt{non-auto}).\\  
\hspace{1em} d) Draw a bounding box: green for \texttt{auto}, red for \texttt{non-auto}.\\  
\hspace{1em} e) Add the confidence label near the bounding box.\\  
\hspace{1em} f) Store the detection in the results list.  

\State \textbf{Step 5:} Return the annotated image and the detection list.  
\end{algorithmic}
\label{algo_detection}

\end{algorithm}

\begin{figure}[t]
    \centering
    \includegraphics[width=0.99\textwidth]{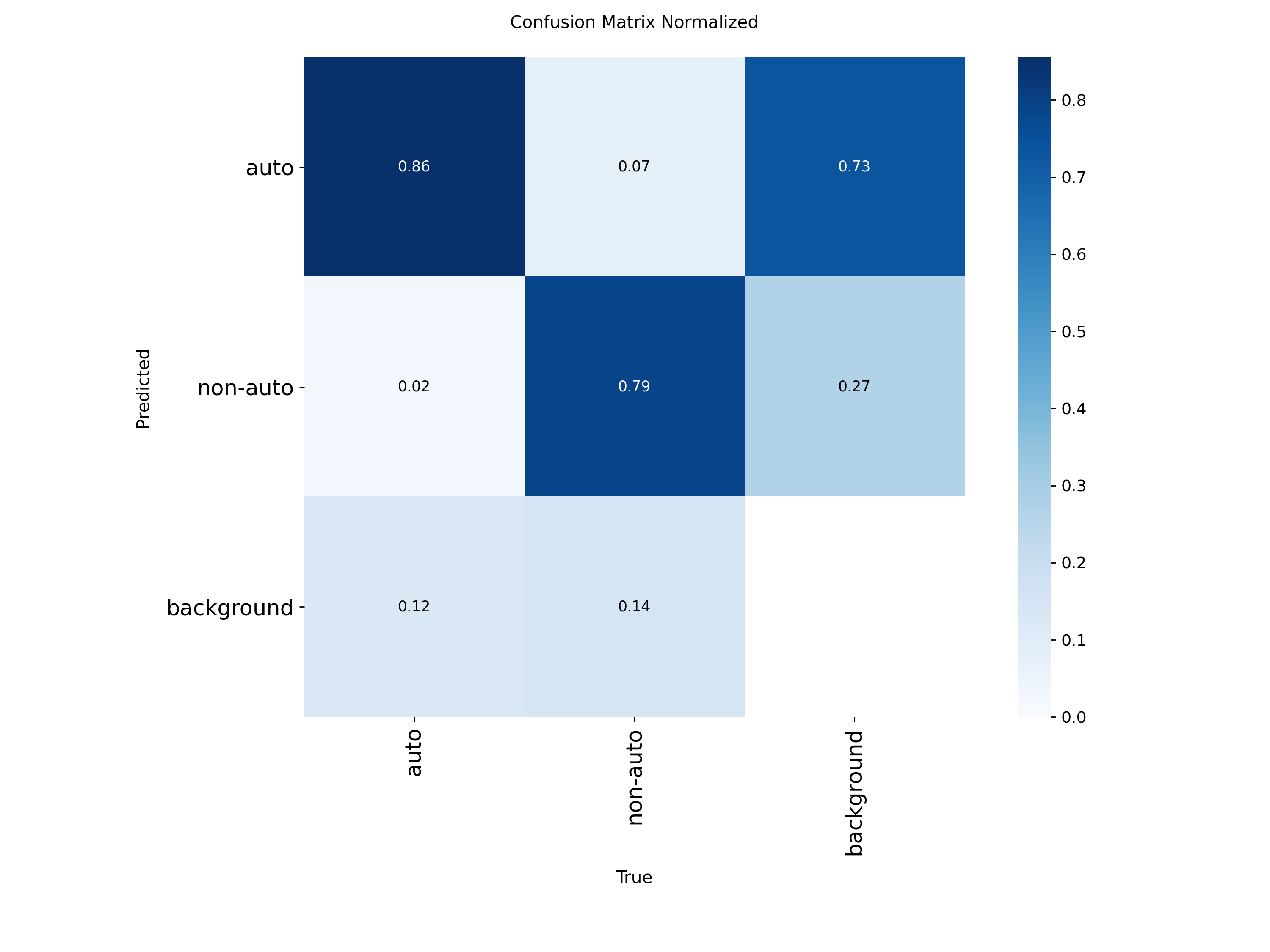}
    \caption{Normalized confusion matrix}
    \label{fig:conf_matrix}
\end{figure}
These evaluations helped determine how well the models would perform in a real-world environment.
In addition to numerical metrics, a qualitative review was conducted on a selected set of images. Even in scenarios with a lot of traffic, our model showed great accuracy in locating auto-rickshaws. Misclassifications and false detections were analyzed separately to identify the potential scope of improvements.

\section{Results and Discussion}\label{sec2}
The model was capable of detecting and localizing auto-rickshaws across a variety of environments, including crowded urban streets with high vehicle density and low-light conditions, such as at night or in shaded regions.

\begin{figure}[t]
    \centering
    % First row
    \begin{subfigure}{0.48\textwidth}
        \centering
        \includegraphics[width=\linewidth]{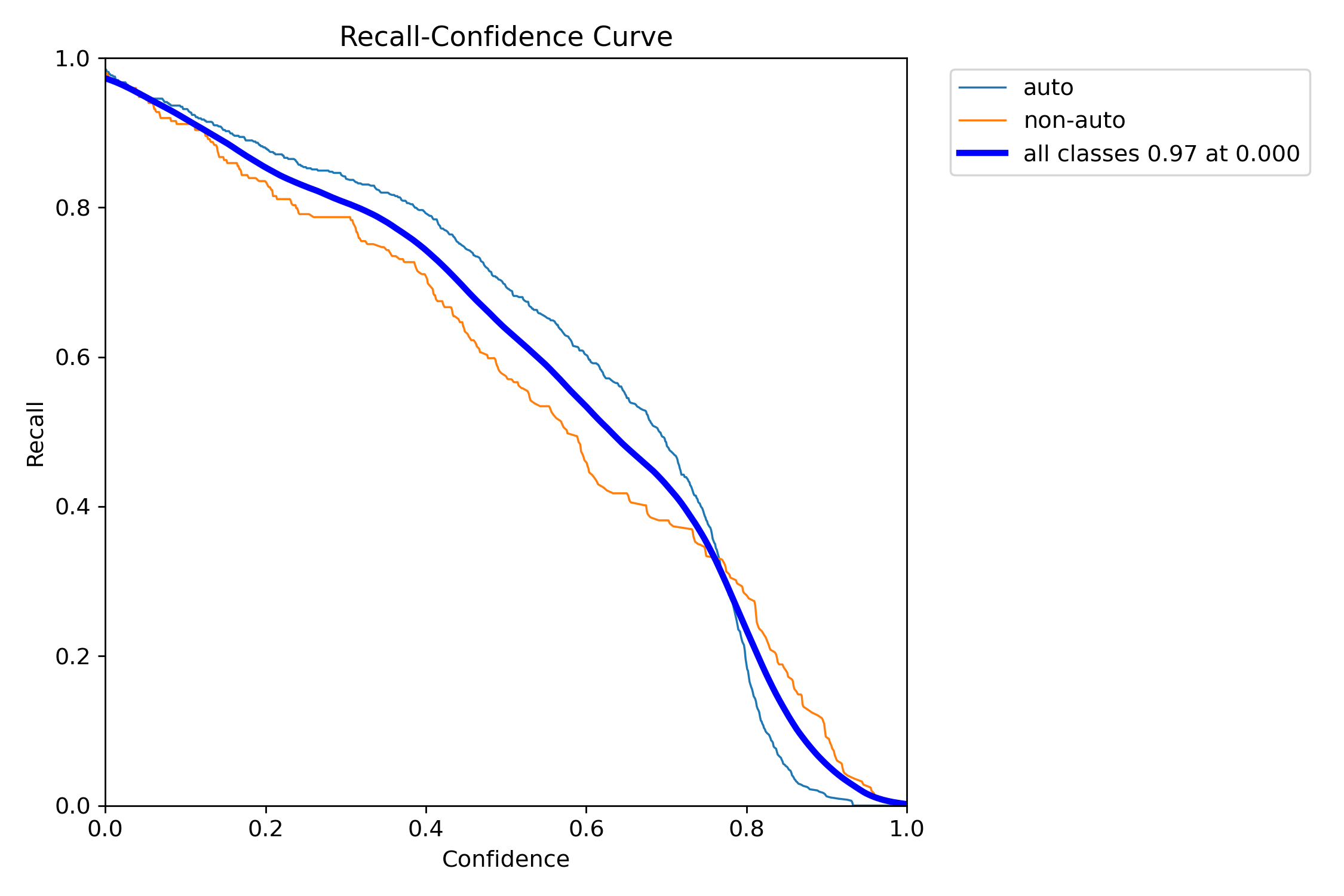}
        \caption{Recall-Confidence Curve}
        \label{fig:recall_curve}
    \end{subfigure}
    \hfill
    \begin{subfigure}{0.48\textwidth}
        \centering
        \includegraphics[width=\linewidth]{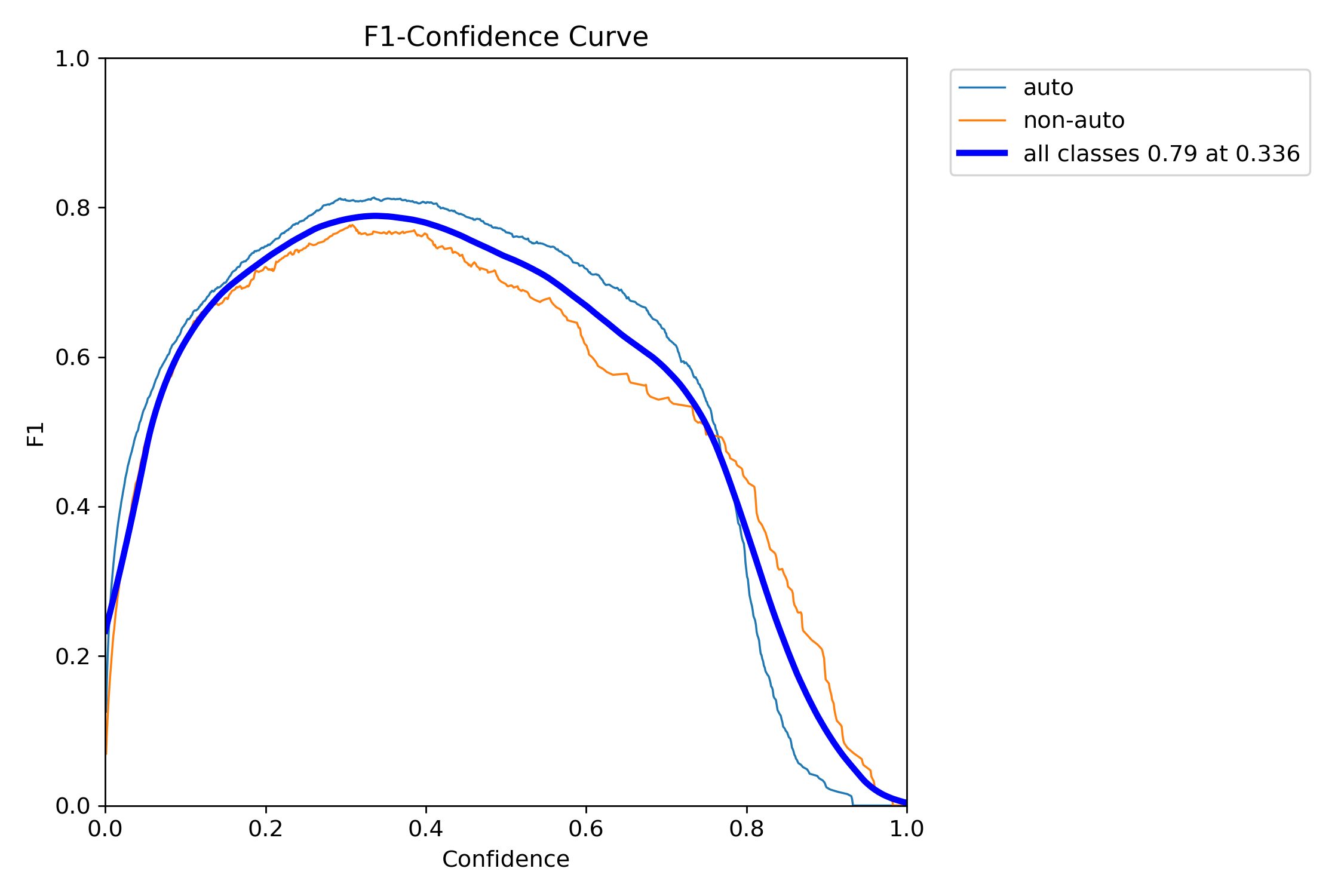}
        \caption{F1-Confidence Curve}
        \label{fig:f1_curve}
    \end{subfigure}
    
    \vspace{1em} % Space between rows
    
    % Second row
    \begin{subfigure}{0.48\textwidth}
        \centering
        \includegraphics[width=\linewidth]{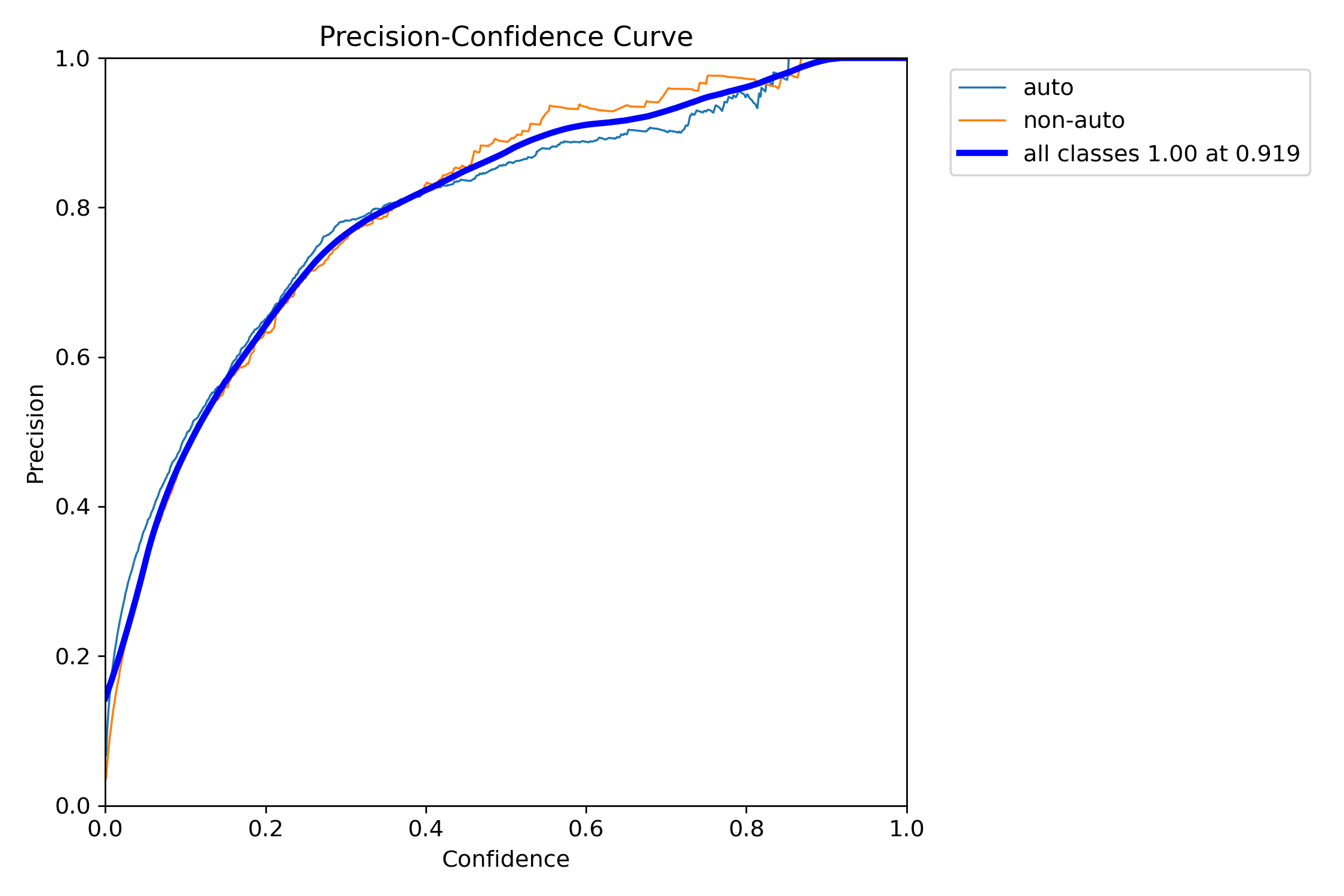}
        \caption{Precision-Confidence Curve}
        \label{fig:precision_curve}
    \end{subfigure}
    \hfill
    \begin{subfigure}{0.48\textwidth}
        \centering
        \includegraphics[width=\linewidth]{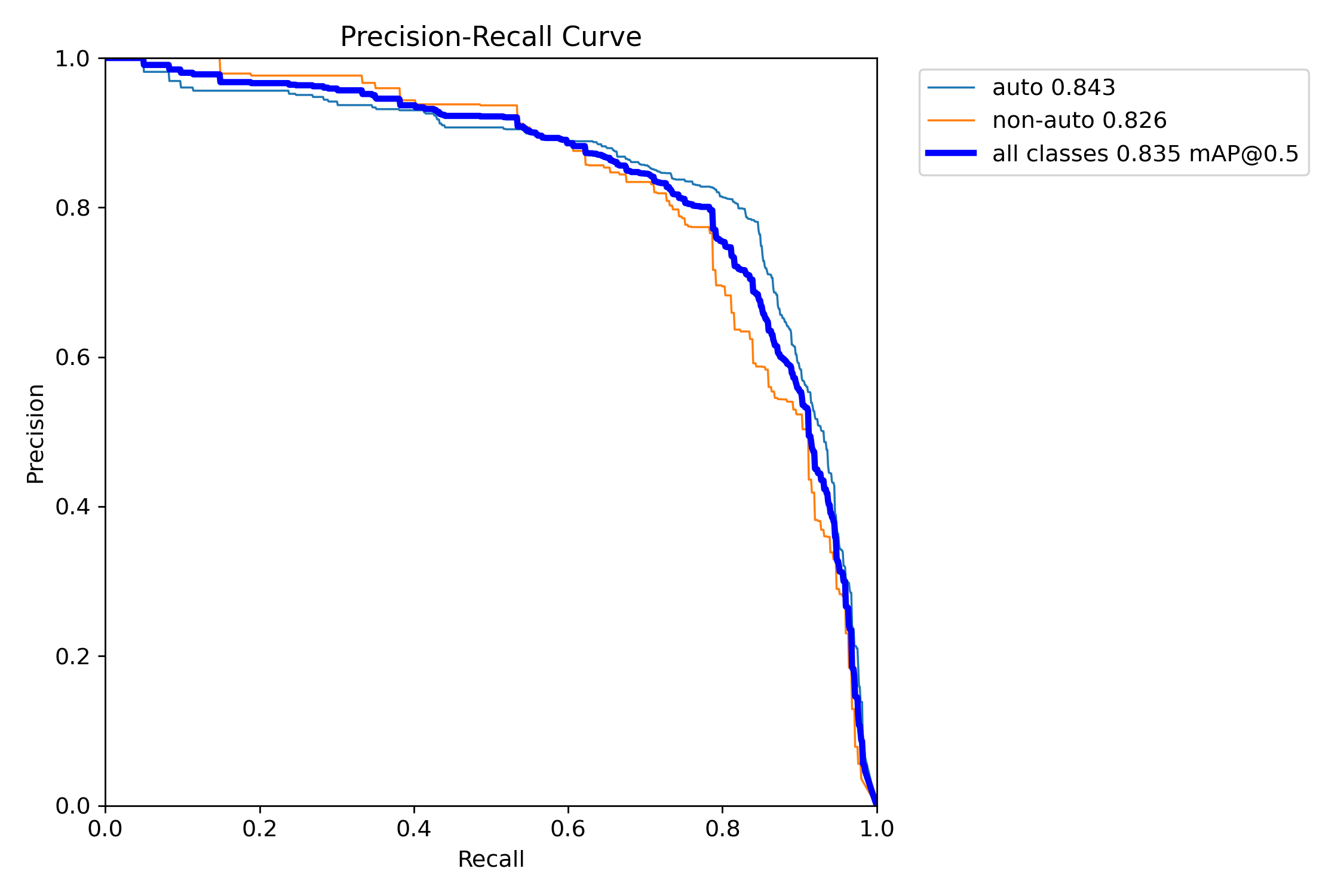}
        \caption{Precision-Recall Curve}
        \label{fig:pr_curve}
    \end{subfigure}
    
    \caption{Performance curves of the model}
    \label{fig:performance_curves}
\end{figure}

Fig. \ref{fig:conf_matrix} and Fig. \ref{fig:performance_curves} represents a qualitative overview of the result of the experiment. Table \ref{result-table} below represents the values of the evaluation metrics of our experiment.

\begin{figure}[H]
    \centering
    
    % First row (3 images)
    \begin{subfigure}{0.32\textwidth}
        \centering
        \includegraphics[width=\linewidth,height=4.2cm]{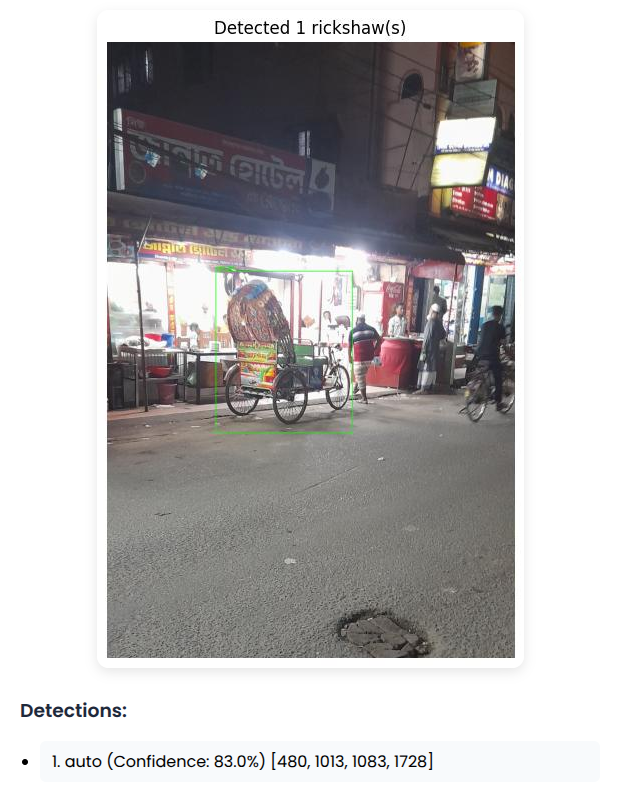}
        \label{fig:recall_curve}
    \end{subfigure}
    \hfill
    \begin{subfigure}{0.32\textwidth}
        \centering
        \includegraphics[width=\linewidth,height=4.2cm]{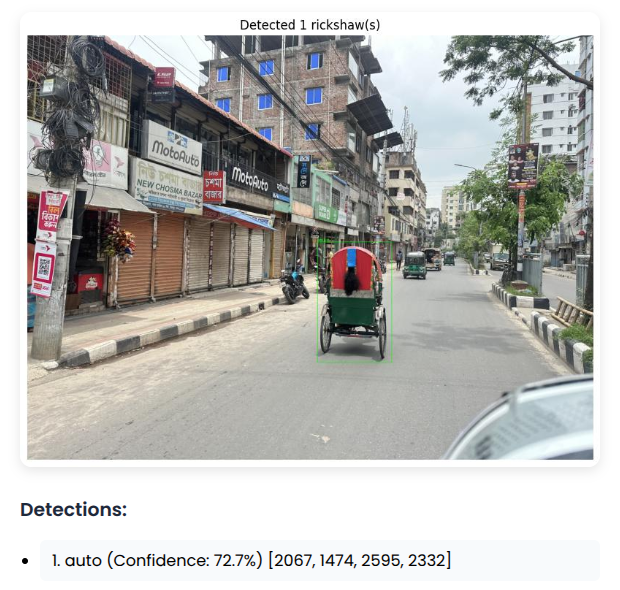}
        \label{fig:f1_curve_1}
    \end{subfigure}
    \hfill
    \begin{subfigure}{0.32\textwidth}
        \centering
        \includegraphics[width=\linewidth,height=4.2cm]{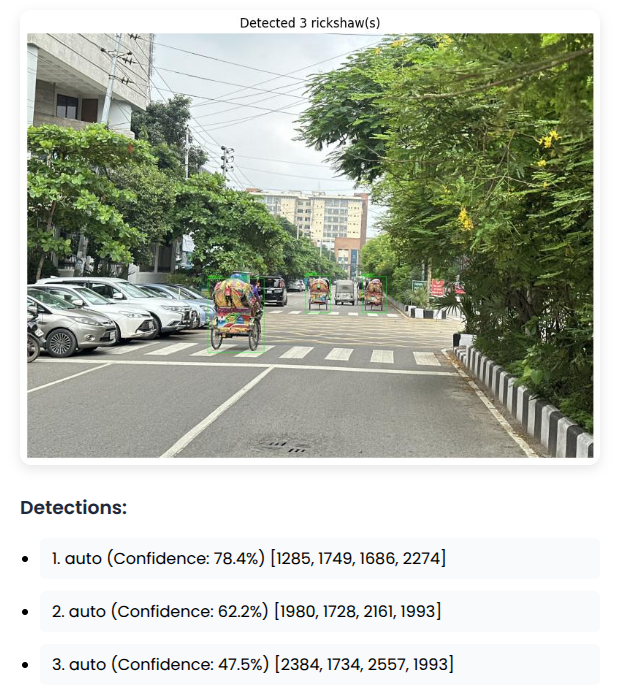}
        \label{fig:precision_curve}
    \end{subfigure}
    
    \vspace{1em} % Space between rows
    
    % Second row (3 images)
    \begin{subfigure}{0.32\textwidth}
        \centering
        \includegraphics[width=\linewidth,height=4.2cm]{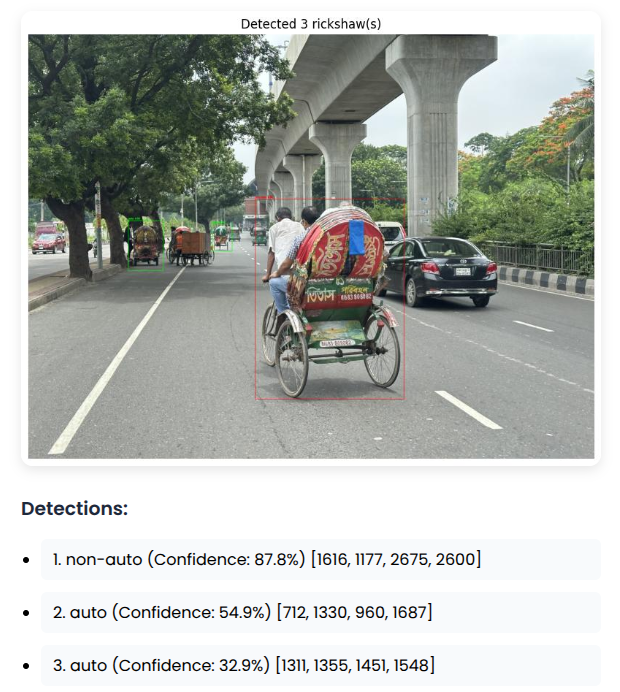}
        \label{fig:pr_curve}
    \end{subfigure}
    \hfill
    \begin{subfigure}{0.32\textwidth}
        \centering
        \includegraphics[width=\linewidth,height=4.2cm]{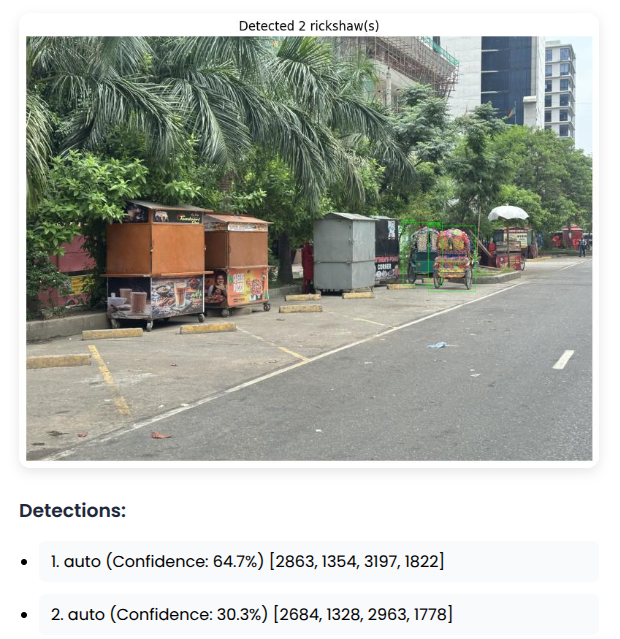}
        \label{fig:f1_curve_2}
    \end{subfigure}
    \hfill
    \begin{subfigure}{0.32\textwidth}
        \centering
        \includegraphics[width=\linewidth,height=4.2cm]{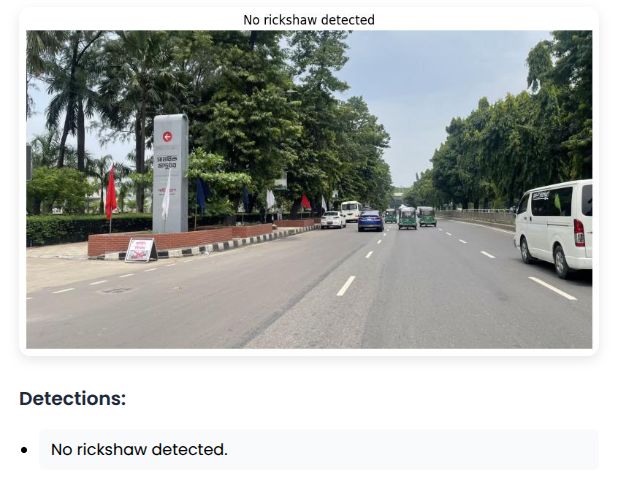}
        \label{fig:f1_curve_3}
    \end{subfigure}
    
    \caption{Sample images of results}
    \label{fig:result_images}
\end{figure}

\begin{table}[h]
\centering
\begin{tabular}{|c|c|c|}
\hline
\textbf{Metric} & \textbf{Value} & \textbf{Percentage(\%)} \\
\hline
Precision (mAP50--95) & 0.55343 & 55.343 \\
\hline
Precision (Binary) & 0.79174 & 79.174 \\
\hline
Recall (Binary) & 0.78798 & 78.798 \\
\hline
mAP50 (Binary) & 0.83447 & 83.447 \\
\hline
\end{tabular}
\caption{Results on validation dataset}
\label{result-table}
\end{table}
These evaluation metrics reflect the overall performance of the model in the validation dataset. After analyzing the results, we can point out the following behavior and detection characteristics of the model.
The trained YOLOv8n model demonstrated accurate detecting capabilities while maintaining a low rate of false positives, even in visually complex urban environments. However, the presence of visually similar vehicles occasionally led to misclassifications, indicating that additional and more diverse training data could further enhance the model’s robustness. The precise implementation of bounding box annotations contributed to a noticeable improvement in the model’s confidence scores, underscoring the importance of accurate labeling in object detection performance. Some sample images of the results of our model are illustrated in Fig. \ref{fig:result_images}.

Based on the general findings of the experiment, the model is scalable and reliable for detecting auto-rickshaws in real-time in urban traffic monitoring systems.

\section{Conclusion}
This project represents the potential of deep learning and computer vision for automated traffic surveillance. Using a custom dataset, we addressed a critical gap in global datasets that often ignore vehicles unique to some particular regions. Using a custom dataset, the model has achieved an mAP50 of 83.447\% and binary precision and recall values above 78\%, demonstrating its effectiveness in handling both dense and sparse traffic scenarios. From the general findings, we can conclude that the model is a scalable and appropriate choice for real-time implementation in traffic surveillance systems. Moreover, the curated dataset provides scope for further study and advancement in the field of region-specific intelligent transportation systems. In the future, we plan to expand our research by connecting the trained model to live traffic feeds to enable real-time detection and monitoring. The scope of the dataset can be expanded to include additional cities and diverse traffic conditions, thereby improving the model’s generalizability. Moreover, incorporating a wider range of vehicle types will enhance classification accuracy in complex traffic environments. Finally, efforts will be made to optimize the model for deployment on low-power edge devices such as Raspberry Pi and NVIDIA Jetson Nano. Through these steps, we aim to scale the system for broader deployment in smart transportation infrastructure in the future.

\vspace{2em}  % adds vertical space (adjust 1em, 2em, etc. as needed)

%\noindent\textbf{Acknowledgements:} We thank Dr. Muhammad Ibrahim, Associate Professor, Department Of Computer Science \& Engineering, University Of Dhaka, for his meticulous guidance throughout the project. 

\section*{Declarations}

\begin{description}
    \item[\textbf{Funding.}] Not applicable.
    \item[\textbf{Conflict of interest/Competing interests.}] The authors declare that they have no competing interests.
    \item[\textbf{Ethics approval and consent to participate.}] Not applicable.
    \item[\textbf{Consent for publication.}] Not applicable.
    \item[\textbf{Data availability.}] The dataset used in this study was collected by the authors. It is not publicly available due to privacy concerns, but it can be shared on a reasonable request.
    \item[\textbf{Availability of materials.}] Not applicable.
    \item[\textbf{Code availability.}] The implementation of the detection model is based on the YOLOv8 framework~\cite{autogyro2023yolov8}.
    \item[\textbf{Author contribution.}] The authors divided the total work among themselves, some collecting images and annotating, some writing code and some writing the manuscript.
\end{description}





\nocite{*}
\bibliography{sn-bibliography}% common bib file

\end{document}